\def\year{2019}\relax
\documentclass[letterpaper]{article} 
\usepackage{aaai19}  
\usepackage{times}  
\usepackage{helvet}  
\usepackage{courier}  
\usepackage{url}  
\usepackage{graphicx}  

\usepackage{amsfonts}
\usepackage{booktabs} 
\usepackage{subfiles} 
\usepackage{multirow}       
\usepackage{listings}       
\usepackage{amsmath}        
\usepackage{url}            
\usepackage{float}          
\usepackage[ruled]{algorithm2e} 
\usepackage{caption}
\usepackage{subcaption}

\begin{document}
%
\title{Global Explanations of Neural Networks \\ \small{Mapping the Landscape of Predictions}}
\author{\small{Mark Ibrahim*, Melissa Louie*, Ceena Modarres*, John Paisley**} \\
    \small{*Center for Machine Learning, Capital One  **Columbia University}
}

\maketitle
\begin{abstract}

A barrier to the wider adoption of neural networks is their lack of interpretability. While local explanation methods exist for one prediction, most global attributions still reduce neural network decisions to a single set of features. In response, we present an approach for generating global attributions called GAM, which explains the landscape of neural network predictions across subpopulations. GAM augments global explanations with the proportion of samples that each attribution best explains and specifies which samples are described by each attribution. Global explanations also have tunable granularity to detect more or fewer subpopulations. We demonstrate that GAM's global explanations 1) yield the known feature importances of simulated data, 2) match feature weights of interpretable statistical models on real data, and 3) are intuitive to practitioners through user studies. With more transparent predictions, GAM can help ensure neural network decisions are generated for the right reasons. 

\end{abstract}

%
%

\maketitle

\section{Introduction}

The present decade has seen the widespread adoption of neural networks across many areas from natural language translation to visual object recognition \cite{lecun2015deep}.
In domains where the stakes are high however, adoption has been limited due to a lack of interpretability. 
The hierarchical non-linear nature of neural networks limits the ability to explain predictions.
This black-box nature exposes the risk of adverse societal consequences such as bias and prejudice.

Researchers have developed several approaches to explain neural network predictions through an observation's saliency or unique identifying factors 
\cite{kindermans2018the}\cite{SimonyanVZ13}\cite{MontavonBBSM15}. 
Another set of techniques produce explanations using attributions, which estimate each feature's importance to a prediction.
Some of these techniques are model agnostic while others attempt to take advantage of a neural network's architecture to generate attributions 
\cite{LundbergL17}\cite{kindermans2018learning} \cite{BinderMBMS16}.
However, these techniques are local, meaning the explanations are limited to a single prediction.

Global attributions can be a powerful tool for interpretability, since they explain feature importance across a population.
Global explanations can aid in model diagnostics, uncover opportunities for feature engineering, and expose possible bias.
Existing global techniques rely on more interpretable surrogate models such as decision trees  or
manipulate the input space to assess global predictive power \cite{frosst2017distilling} \cite{yang2018global} \cite{lakkaraju2016interpretable}.
While these techniques produce an interpretable set of rules, they can fail to capture the non-linear feature interactions learned by neural networks.
Submodular-Pick LIME (SP-LIME) is another technique based on summarizing local attributions to better reflect learned interactions.
Although SP-LIME maximizes coverage of local explanations, it does not describe the proportion or subpopulations best explained by each global attribution \cite{RibeiroSG16}.

We propose a global attribution method called GAM (Global Attribution Mapping) that's able to explain non-linear representations learned by neural networks across subpopulations.
GAM groups similar local feature importances to form human-interpretable global attributions. 
Each global attribution best explains a particular subpopulation, augmenting global explanations with the proportion and specific samples they best explain.
In addition, GAM provides tunable granularity to capture more or fewer subpopulations for global explanations.

\section{Background: Attribution Techniques}

GAM builds on local attribution techniques. Here we consider three techniques that are rooted in a strong theoretical foundation, well cited, and relatively easy to implement: Locally Interpretable Model-Agnostic Explanations (LIME), Integrated Gradients, and Deep Learning Important FeaTures (DeepLIFT) \cite{RibeiroSG16} \cite{SundararajanTY17} \cite{ShrikumarGK17}. 

\paragraph{Locally Interpretable Model-Agnostic Explanations (LIME)}
LIME produces interpretable representations of complicated models by optimizing two metrics: interpretability and local fidelity.
In practice, an input is perturbed by sampling in a local neighborhood to fit a simpler linear model. 
The now explainable linear model's weights can be used to interpret a particular model prediction.

There is a global extension of LIME called Submodule pick LIME (SP-LIME).
SP-LIME reduces many local attributions into a smaller global set by selecting the least redundant set of local attributions.
Formally, SP-LIME maximizes coverage defined as the total importance of the features in the selected global set.

\paragraph{Integrated Gradients}
Sundararajan et. al developed an approach that satisfies two proposed axioms for attributions. First, sensitivity states that for any two inputs that differ in only a single feature and have different predictions, the attribution should be non-zero. Second, implementation invariance states that two networks that have equal outputs for all possible inputs should also have the same attributions. The authors develop an approach that takes the path integral of the gradient for a neutral reference input to determine the feature dimension with the greatest activation.
An attribution vector for a particular observation (local) is produced as a result.

\paragraph{Deep Learning Important FeaTures (DeepLIFT)}
Much like Integrated Gradients, DeepLIFT seeks to explain the difference in output between an input and a neutral reference. An attribution score is calculated by constructing 
a function analogous to a partial derivative that's used along with the chain rule to trace the change in the output layer relative to the input.

\section{GAM: Global Attribution Mapping}
\label{method}

We treat each local attribution as a ranking of features.
In contrast to domains such as speech or image recognition where individual inputs are not human-interpretable, here we focus on situations where features have well defined semantics 
\cite{datta2016algorithmic}. We then group similar attributions to form global human-interpretable explanations across subpopulations.

\subsection{Attributions as Rankings}
Attributions contain more than the content encoded by treating each as a vector in $\mathbb{R}^n$:
an attribution ranks and weighs each feature's association with a particular prediction. 
Each element in an attribution vector answers the question: "how important is this feature for a particular prediction?"
Furthermore, attributions are conjoined rankings, because every feature appears in every attribution along the same dimension.
Therefore to incorporate both the rank and weight encoded in an attribution, we treat attributions as \textit{weighted conjoined rankings}.
With this treatment, we summarize the content of many local attributions by comparing their similarities to form global attributions.

\subsection{Rank Distances}

We represent each attribution as a rank vector $\sigma$. 
The unweighted rank of feature $i$ is $\sigma(i)$ and the weighted rank is $\sigma_w(i)$.

First, to ensure distances appropriately reflect similarity among vectors---as opposed to anomalies in the scale of the 
original inputs---we normalize each local attribution vector $\sigma_w$ by 

$$ |\sigma_w| \circ \frac{1}{\sum_i |\sigma_w(i)|}, $$

where $\sigma_w$ is the weighted attribution vector and $\circ$ is the Hadamard product.

By transforming the attribution into normalized percentages, we consider only feature importance, not whether
a feature contributes positively or negatively to a particular target. 
This enables us to measure similarity among normalized attributions as weighted rankings 
of feature importance.  

Next we consider two options for comparing attributions: Kendall's Tau and Spearman's Rho squared rank distances.
The first option for comparing attributions is weighted Kendall's Tau rank distance, an extension of Kendall's Tau rank correlation as defined by
Lee and Yu 2010 \cite{Lee2010}. 

For a second attribution vector $\pi$, we can define the weighted Kendall's Tau distance between $\pi$ and $\sigma$ as

\begin{equation} \label{eq:1}
\sum_{i < j} w_i w_j I\{(\pi(i) - \pi(j))(\sigma(i) - \sigma(j)) < 0\}, 
\end{equation}

where $w_i = \pi_w(i) \sigma_w(i)$ and $I$ is the indicator function.

A feature's weight, $w_i$, in this context is the product of the weights in each ranking. 
Two rankings containing all elements in the same order (possibly different weights) have zero distance to imply the same feature importance. 
Two rankings with at least one element in a different order have a distance proportional to the out-of-order elements' weights.
When a feature is ranked above or below in one ranking compared to the other, it is penalized by the product of its weights in the rankings.
The total distance between two rankings is then the sum of the penalty for feature appearing in a different order.
The more features that appear in a different order, the greater the distance between the rankings.
Furthermore, the weighted Kendall's Tau distance described satisfies
symmetry, positive definiteness, and subadditivity (see Section 2 of \cite{LeeMWD}) meaning it defines a proper distance metric 
that we can use for clustering rankings.

A second option for comparing attributions is weighted Spearman's Rho squared rank distance, as defined by Shieh et al. 
\cite{Shieh2000},

\begin{equation} \label{eq:2}
    \sum_{i = 1}^k \pi_w(i) \sigma_w(i) (\pi(i) - \sigma(i))^2.
\end{equation}

Weighted Spearman's Rho squared rank distance as defined compares attributions as weighted conjoined rankings much in the same way as 
weighted Kendall's Tau rank distance, but with a 
runtime of $\mathcal{O}(n\log{}n)$. 
Weighted Spearman's Rho squared distance also defines a distance metric \cite{LeeMWD}.

Kendall's Tau rank distance does have computational and mathematical limitations.
First, Kendall's Tau has a quadratic runtime, which can be computationally expensive as the number of feature and/or samples grows 
\cite{Knight1966}.
Second, Kendall's Tau tends to produce smaller values compared to Spearman's Rho, making differences between attributions less pronounced
\cite{FREDRICKS2007}.

Therefore, in situations where the number of features and/or samples is large, Spearman's Rho squared is a more effective alternative
from a computational perspective and due to its ability to make differences among rankings more pronounced.
Despite the advantages of Spearman's Rho squared rank distance, 
Kendall's Tau remains an appropriate choice when computational efficiency is less important due to its desirable statistical properties. 
Kendall's Tau as a correlation metric has faster asymptotic convergence with respect to bias and variance \cite{XU2013}.
Having defined metrics with which to compare attributions, we now turn our attention to identifying similar attributions.

\subsection{Grouping Similar Local Attributions}
\label{clusteringAttributions}

Clustering algorithms can detect global patterns in a dataset by grouping similar data points into a cluster. 
We apply this idea in the context of local attribution vectors by transforming many local attributions into a few global attributions.  
We use weighted Kendall's Tau or Spearman's Rho squared rank distances to measure similarity among attributions.
We adapt K-medoids to use rank distance rather than Manhattan distance to group similar attributions
\cite{Park2009}.
We then use the resulting medoids to succinctly summarize global attribution patterns.

The clustering algorithm identifies the K local attributions (medoids) minimizing the pairwise dissimilarity within a cluster relative to the medoid.
We then use the medoid to summarize the pattern detected for each cluster. 
The algorithm works by initializing K local attributions (medoids) $m_1, ..., m_K$ selected randomly from the full set of local attributions. 
The local attributions are reassigned until convergence or for a sufficiently large number of iterations, each taking $\mathcal{O}(n(n-K)^2)$.
The choice of medoids at each iteration is updated by
\begin{enumerate}

\item determining cluster membership for each local attribution $\sigma$ via the nearest medoid,
	$$  \sigma \in C_a  \textrm{ if } \text{rankDist}(\sigma, m_a) \leq \text{rankDist}(\sigma, m_j)\  \forall j \in [1, K], $$
\item updating each cluster's medoid by minimizing pairwise dissimilarity within a cluster,
	$$ m_a = \sigma_l = argmin_{\sigma_l \in C_a}  \sum_{\sigma_j \in C_a} \text{rankDist}(\sigma_j, \sigma_l). $$

\end{enumerate}

The advantage this clustering approach over other algorithms such as K-means is that it allows us to compare
attributions as conjoined weighted rankings rather than simply vectors in $\mathbb{R}^N$. 
K-means, for example, treats attributions as vectors by minimizing inter-cluster variance in Euclidean space, 
ignoring the content encoded in the rank and weights of each feature.
Our approach, on the other hand, takes advantage of both the rank and weight information in local attributions during clustering. 
Each cluster specifies a subpopulation best explained by its medoid.

\subsection{Generating Global Attributions}

Each of GAM's global explanations yields the most centrally located vector of feature importances for a subpopulation.
The global attribution landscape described by GAM is then the collection of global explanations for each subpopulation.

Finally, we form GAM's global explanations across subpopulations by grouping similar normalized attributions.

\IncMargin{1em}
\begin{algorithm}[t]
\SetAlgoNoLine
\KwIn{local attributions}
\KwOut{medoids and corresponding members}
\BlankLine
    \tcc{1. Normalize the set of local attributions}
    \ForEach{ local attribution}{
        normalized = abs(attribution) / sum(abs(attribution))}

\BlankLine
\BlankLine
    \tcc{ 2. Compute pair-wise rank distance matrix}
distances = $[]$
\BlankLine
    \ForEach{attribution1 in normalizedAttributions}{
        \ForEach{attribution2 in normalizedAttributions}{
            distances += rankDistance(attribution1, attribution2)}}

\BlankLine
\BlankLine
    \tcc{3. Cluster Attributions}
initialMedoids = random.choice(attributions)

\BlankLine
    \For{x iterations}{
        \ForEach{cluster}{
            \ForEach{attribution in cluster}{
                tempMedoid = attribution\;
                cost = sum(distance(attribution, tempMedoid))\;
                reassign medoid to attribution minimizing cost\;}
        update cluster membership by assigning to closest medoid}}

\caption{Generating Global Attributions (GAM)}
\end{algorithm}

We can gauge the explanatory power of each global attribution by the size of its subpopulation.
Since GAM allows us to trace a global explanation to individual samples, we can also examine summary statistics for each subpopulation.
In addition, by adjusting the cluster size $K$, 
we can tune the granularity of global explanations to capture more or fewer subpopulations.
As a starting point, $K$ can be chosen based the Silhouette score, which compares similarity within and across clusters
\cite{rousseeuw1987silhouettes}.

In contrast to global explanations via surrogate models, which produce a single global explanation, 
GAM identifies differences in explanations among subpopulations.
Furthermore, the ability to trace global explanations to individual samples, supplements 
global explanations of techniques such as SP-LIME with information about subpopulations. 
\section{Validating Our Proposed Global Attribution Method}

Although there has been significant research in the validation of clustering algorithms, interpreting clusters involves qualitative analysis and domain expertise. 
Moreover, research in interpretability inherently faces the challenge of effective and reliable validation \cite{doshi2017towards}. 
Attributions have no baseline truth. 
The task of identifying an appropriate validation methodology for a proposed approach is an open research question. 
In this paper, we validate the methodology in three ways.

The first validates GAM against two synthetically generated datasets with known feature importances. 
The second compares the global attributions generated by GAM against the feature weights of more interpretable statistical models.
The third assesses the practical value of GAM in real-world settings with user studies.
We also demonstrate the use of Kendall's Tau and Spearman's Rho rank distances as well as 
several local attribution techniques.

\subsection{Global Attributions Applied to Synthetic Dataset}

We design a simple synthetic dataset to yield known feature importances against which to validate GAM's global attributions.
The dataset consists of a binary label, class 1 or 2, and two features, A and B. 
We aim to generate one group for which only feature A 
is predictive and another group for which only feature B is predictive.

For the group where only feature A is predictive, we sample balanced classes from uniform distributions where 

\begin{equation}
    \text{feature A} = \left\{
     \begin{array}{lr}
         U[0, 1] & \text{if class 1} \\
         U[1, 2] & \text{if class 2}
     \end{array}
    \right\}. 
\end{equation}

To ensure feature B has no predictive power for this group, we draw from a uniform distribution over $[0, 2]$.

Since values for feature A in $[0, 1]$ mark class 1 and $[1, 2]$ mark class 2 samples, feature A has full predictive power.
Feature B on the other hand is sampled uniformly over the same range for both classes, yielding no predictive power. 

Conversely, to produce a group where only feature B is predictive, we 
sample balanced classes where

\begin{equation}
    \text{feature B} = \left\{
     \begin{array}{lr}
         U[3, 4] & \text{if class 1} \\
         U[4, 5] & \text{if class 2}
     \end{array}
    \right\}. 
\end{equation}

To ensure feature A has no predictive power for this group, we draw from a uniform distribution over $[3, 5]$.

We then conduct two experiments:
1. using a balanced set where feature A is predictive in 50\% of cases and 
2. using an unbalanced set where feature A is predictive in 75\% of cases.
We train a single layer feed forward neural network on a balanced dataset of 10k samples.
We use a ReLU activation with a four node hidden layer and a sigmoid output. 
We hold out 2000 test samples and achieve a validation accuracy of 99\%.

In order to generate global attributions for the 2000 balanced test samples, we first generate local attributions using LIME. 
We then generate global attributions using GAM across several values of $K$.
By computing the Silhouette Coefficient score, we find $K = 2$ optimizes the similarity of attributions within the subpopulations.
We obtain the corresponding size of the subpopulations and two explanation medoids summarizing feature importances (see Table \ref{tab:synthetic}).

\begin{table}[h]
\centering
\caption{Global Attributions for Synthetic Data}
\label{tab:synthetic}
\begin{tabular}{|c|c|}
\hline
    \multicolumn{2}{| c |}{Balanced} \\
\hline
Explanations & Subpopulation Size  \\

\hline
$\begin{bmatrix} \text{feature A: } 0.93 \\ \text{feature B: } 0.07 \end{bmatrix}$ & 
    953   \\
\hline
        $\begin{bmatrix} \text{feature A: }  0.05 \\ \text{feature B: }  0.95 \end{bmatrix}$ 
            &  1047\\
\hline
\end{tabular}
\end{table}

\begin{table}[h]
\centering
\caption{Global Attributions for Unbalanced Synthetic Data}
\label{tab:syntheticUnbalanced}
\begin{tabular}{|c|c|}
\hline
 \multicolumn{2}{| c |}{Unbalanced} \\
\hline
Explanations & Subpopulation Size \\

\hline
    $\begin{bmatrix} \text{feature A: }  0.94 \\ \text{feature B: }  0.06 \end{bmatrix}$ 
        & 1436 \\
\hline
        $\begin{bmatrix} \text{feature A: }  0.13 \\ \text{feature B: }  0.87 \end{bmatrix}$ 
            &  564 \\
\hline
\end{tabular}
\end{table}

For the balanced experiment, the global explanations and subpopulation proportions match expectations. 
The first subpopulation contains nearly half the samples and assigns a high importance weight to feature A. 
The second subpopulation contains the remaining half of samples and assigns a high importance weight to feature B.

For the unbalanced experiment, the global attributions contain two subpopulations with the expected proportions (see Table \ref{tab:syntheticUnbalanced}).
The first assigns a high attribution to feature A and contains 72\% of samples.
The second assigns a high attribution to feature B and contains 28\% of samples, 
reflecting the feature importances inherent to the synthetic dataset. 
See Appendix for discussion on DeepLIFT and Integrated Gradients results.

Overall, the global attributions produced by GAM match the known feature importance and proportions of both synthetic datasets.

\subsection{Global Attributions Applied to the Mushroom Dataset}

Next, we apply GAM to a dataset with a larger number of categorical features: The Audubon Society's mushroom dataset
\cite{Dua2017}.
The task is to classify mushroom samples as poisonous or not poisonous using 23 categorical features such as cap-shape, habitat, odor and so on. 
Here we highlight the use of gradient-based neural network attribution techniques (DeepLIFT and Integrated Gradients) to generate local attributions
and compare the results against LIME.
We also choose weighted Spearman's Rho squared rank distance to compare attributions for computational efficiency given the larger number of features and samples.

For the classification task, we train a two hidden-layer feed-foward neural network on 4784 samples (witholding 2392 for validation) 
based on the architecture proposed 
in the Kaggle Kernel \cite{Kaggle}.
We one-hot encode the categorical features into a 128-unit input layer. 
The network optimizes binary cross entropy with dropout using sigmoid activation functions.
We obtain a validation accuracy of approximately 99\%. This allows us to produce more accurate local attributions by minimizing
inaccuracies due to model error.

\begin{figure}
    \includegraphics[width=\linewidth]{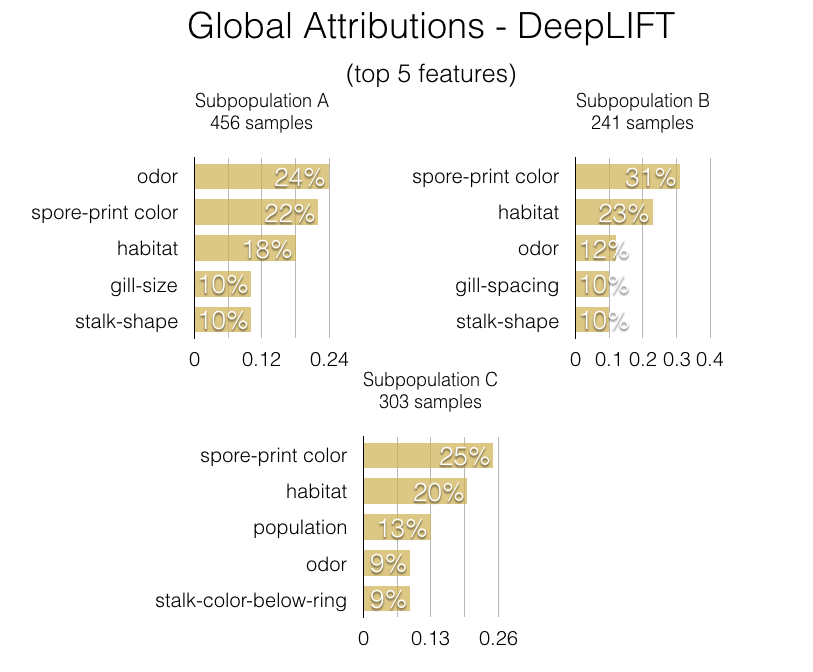}
    \caption{DeepLIFT- Five most important features for each global attribution explanation and the corresponding number of samples 
    in each subpopulation.}
    \label{fig:deepliftMushrooms}
\end{figure}

We apply GAM on three sets of local attributions on 1000 samples using DeepLIFT, Integrated Gradients, and LIME 
\cite{ancona2018towards}. We identify three global explanations based on each technique. We note the top features did vary slightly in weight and rank depending on our choice of baseline (for DeepLIFT/Integrated Gradients). Here we choose the baseline yielding a nearly uniform classification output.

The global attributions based on DeepLIFT produce three subpopulations with similar top features (see Figure \ref{fig:deepliftMushrooms}).
Odor, habitat, and spore-print color appear among the top five features across all three subpopulations. 
Each subpopulation also surfaces distinctions in the relative feature importances.
Spore-print color appears to be the most important features for 
54\% of samples, while odor is most important for the remaining subpopulation. 
Gill-size also appears to be relatively more important for subpopulation A, whereas gill-spacing is more so for subpopulation B.
Despite slight variations in rankings, on average 8 out of the top 10 features agree across all three subpopulations, implying overall concordance in the surfaced top features (see Appendix for full attributions).

\begin{figure}
    \includegraphics[width=\linewidth]{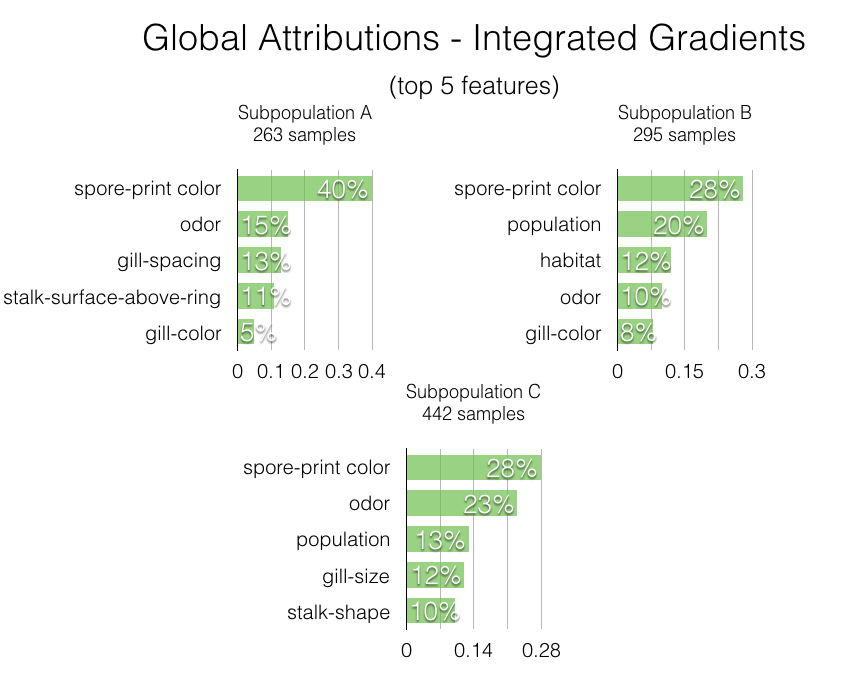}
    \caption{Integrated Gradients - Global Attributions}
    \label{fig:igMushrooms}
\end{figure}

The global attributions based on Integrated Gradients also produce explanations ranking odor and spore-print-color among the top five features. 
There appears to also be strong overlap among other top features such as habitat, population, and stalk-shape 
(see Figure \ref{fig:igMushrooms}).
Nevertheless, each subpopulation also surfaces nuances in the top features for each group. 
For example, population is among the top 5 features for 74\% of samples, whereas it's relatively less important for the remaining samples. 
Across all three subpopulations however, on average 7 of the top 10 features agree, once again implying concordance in the top features.

\begin{figure}
    \includegraphics[width=\linewidth]{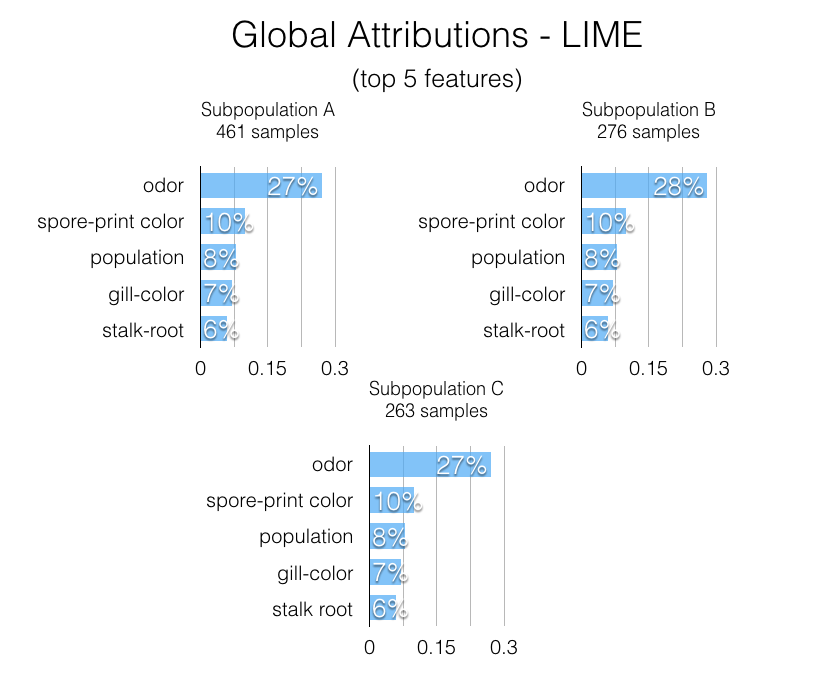}
    \caption{LIME - Global Attributions}
\label{fig:limeMushrooms}
\end{figure}

In addition, we find both gradient-based global attributions overlap in the top features for global attributions based on LIME.
Global attributions for LIME also rank odor, spore-print-color, population, and gill-color among the top five features.
The three subpopulations based on LIME produced similar top five features 
(see Figure \ref{fig:limeMushrooms}). 
Despite slight variation in rankings, all three local techniques surface similar most important features.

\begin{figure}
    \includegraphics[width=\linewidth]{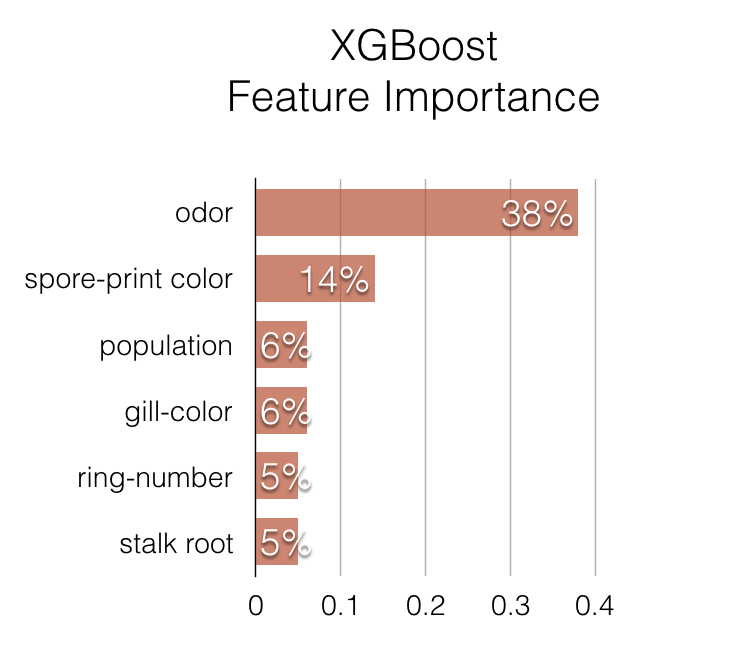}
\caption{Feature importances for XGBoost as measured by each feature's F1 score.}
\label{fig:xgboostMushrooms}
\end{figure}

Finally, we validate global attributions across all three local techniques against the known feature importances of a gradient boosting tree classifier.
We use the XGBoost classifier implementation and assign feature importances using each feature's F1 score \cite{Chen2016}.
We find on average 7 out of the top 10 features in the global attributions across all three techniques match the top features of XGBoost.
Odor, spore-print, population, gill-color, and stalk root appear among the most important features (see Figure \ref{fig:xgboostMushrooms}).
Furthermore, the weights of the top features of XGBoost match closely the weights of global attributions based on LIME.
Although global XGBoost feature importances corroborate the most important features surfaced, 
we qualify these findings by noting that two equally performing models can learn a different set of feature interactions.
Nevertheless, in this case the global attributions' similarity to known global feature importances across multiple local techniques suggests
the explanations generated effectively uncover associations among features and the classification task.

We found similar agreement in the top features using GAM on the Iris dataset with global explanations matching the weights of a logistic classifier (see Appendix: GAM Applied to the Iris Dataset).

\subsection{User Studies}
We conduct several user studies on FICO Home Equity Line Credit applications to assess the practical value of GAM in real-world settings. 
The dataset provides information on credit candidate applications and risk outcome characterized by 90-day payment delinquencies. 

First, we simulated a scenario where a practitioner would use GAM to conduct feature selection. We selected seven predictive features and added in three additional uniformly distributed noise features. We dropped all feature names, 
trained a neural network to make credit decisions, produced a GAM model explanation, and 
presented the explanation to 55 Amazon Mechanical Turk users. 
We provided users with GAM explanations in the form of a bar chart and asked users to remove the three least important features.
In 94\% of cases, respondents selected the features with the lowest importance corresponding to the random noise features.

Second, we polled credit modeling experts to assess whether GAM's results are as intuitive as those of the popular SP-LIME global attribution method.
We present both explanations to a group of 
55 credit modeling practitioners
and asked which explanation allows them to more intuitively improve the model (see Appendix).
We found respondents had nearly equal preference among the three categories: GAM, SP-LIME, and no preference.

Our sample sizes are relatively small, but the results suggest that GAM has real world practical value in machine learning tasks like feature selection and 
produces results that are intuitive to human practitioners.

\section{Conclusion}

In this paper, we develop GAM, a methodology for generating global attributions to supplement existing interpretability techniques for neural networks. 
GAM's global explanations describe the non-linear representations learned by neural networks.
GAM also provides tunable subpopulation granularity along with the ability to trace global explanations to specific samples.
We demonstrate on both real and synthetic datasets that GAM illuminates global explanation patterns across learned subpopulations.
We validate that global attributions produced by GAM match known feature importances and are insightful to humans through user studies.
With explanations across subpopulations, neural network predictions are more transparent.
A possible next step is to explore how global interpretability can help ensure fairness in AI algorithms.

\section{Acknowledgements}
Thank you to Paul Zeng for his contribution to the implementation and valuable feedback.
An implementation for GAM is available at https://github.com/capitalone/global-attribution-mapping . 

\bibliographystyle{plain}

\begin{thebibliography}{}

\end{thebibliography}


\begin{thebibliography}{10}

\bibitem[\protect\citeauthoryear{Ancona \bgroup et al\mbox.\egroup
  }{2018}]{ancona2018towards}
Ancona, M.; Ceolini, E.; Öztireli, C.; and Gross, M.
\newblock 2018.
\newblock Towards better understanding of gradient-based attribution methods
  for deep neural networks.
\newblock In {\em International Conference on Learning Representations}.

\bibitem[\protect\citeauthoryear{Binder \bgroup et al\mbox.\egroup
  }{2016}]{BinderMBMS16}
Binder, A.; Montavon, G.; Bach, S.; M{\"{u}}ller, K.; and Samek, W.
\newblock 2016.
\newblock Layer-wise relevance propagation for neural networks with local
  renormalization layers.
\newblock {\em CoRR} abs/1604.00825.

\bibitem[\protect\citeauthoryear{Chen and Guestrin}{2016}]{Chen2016}
Chen, T., and Guestrin, C.
\newblock 2016.
\newblock Xgboost: A scalable tree boosting system.
\newblock In {\em Proceedings of the 22Nd ACM SIGKDD International Conference
  on Knowledge Discovery and Data Mining}, KDD '16,  785--794.
\newblock New York, NY, USA: ACM.

\bibitem[\protect\citeauthoryear{Datta, Sen, and
  Zick}{2016}]{datta2016algorithmic}
Datta, A.; Sen, S.; and Zick, Y.
\newblock 2016.
\newblock Algorithmic transparency via quantitative input influence: Theory and
  experiments with learning systems.
\newblock In {\em Security and Privacy (SP), 2016 IEEE Symposium on},
  598--617.
\newblock IEEE.

\bibitem[\protect\citeauthoryear{Dheeru and Karra~Taniskidou}{2017}]{Dua2017}
Dheeru, D., and Karra~Taniskidou, E.
\newblock 2017.
\newblock {UCI} machine learning repository: Mushroom data set.

\bibitem[\protect\citeauthoryear{Doshi-Velez and Kim}{2017}]{doshi2017towards}
Doshi-Velez, F., and Kim, B.
\newblock 2017.
\newblock Towards a rigorous science of interpretable machine learning.
\newblock {\em arXiv preprint arXiv:1702.08608}.

\bibitem[\protect\citeauthoryear{Fredricks and Nelsen}{2007}]{FREDRICKS2007}
Fredricks, G.~A., and Nelsen, R.~B.
\newblock 2007.
\newblock On the relationship between spearman's rho and kendall's tau for
  pairs of continuous random variables.
\newblock {\em Journal of Statistical Planning and Inference} 137(7):2143 --
  2150.

\bibitem[\protect\citeauthoryear{Frosst and
  Hinton}{2017}]{frosst2017distilling}
Frosst, N., and Hinton, G.
\newblock 2017.
\newblock Distilling a neural network into a soft decision tree.
\newblock {\em arXiv preprint arXiv:1711.09784}.

\bibitem[\protect\citeauthoryear{Kim}{2018}]{Kaggle}
Kim, Y.
\newblock 2018.
\newblock Kaggle: Modeling on mushroom dataset.

\bibitem[\protect\citeauthoryear{Kindermans \bgroup et al\mbox.\egroup
  }{2018a}]{kindermans2018the}
Kindermans, P.-J.; Hooker, S.; Adebayo, J.; Schutt, K.~T.; Alber, M.; Dahne,
  S.; Erhan, D.; and Kim, B.
\newblock 2018a.
\newblock The (un)reliability of saliency methods.

\bibitem[\protect\citeauthoryear{Kindermans \bgroup et al\mbox.\egroup
  }{2018b}]{kindermans2018learning}
Kindermans, P.-J.; Schutt, K.~T.; Alber, M.; Muller, K.-R.; Erhan, D.; Kim, B.;
  and Dahne, S.
\newblock 2018b.
\newblock Learning how to explain neural networks: Patternnet and
  patternattribution.

\bibitem[\protect\citeauthoryear{Knight}{1966}]{Knight1966}
Knight, W.~R.
\newblock 1966.
\newblock A computer method for calculating kendall's tau with ungrouped data.
\newblock {\em Journal of the American Statistical Association}
  61(314):436--439.

\bibitem[\protect\citeauthoryear{Lakkaraju, Bach, and
  Leskovec}{2016}]{lakkaraju2016interpretable}
Lakkaraju, H.; Bach, S.~H.; and Leskovec, J.
\newblock 2016.
\newblock Interpretable decision sets: A joint framework for description and
  prediction.
\newblock In {\em Proceedings of the 22nd ACM SIGKDD international conference
  on knowledge discovery and data mining},  1675--1684.
\newblock ACM.

\bibitem[\protect\citeauthoryear{LeCun, Bengio, and
  Hinton}{2015}]{lecun2015deep}
LeCun, Y.; Bengio, Y.; and Hinton, G.
\newblock 2015.
\newblock Deep learning.
\newblock {\em nature} 521(7553):436.

\bibitem[\protect\citeauthoryear{Lee and Yu}{2010}]{Lee2010}
Lee, P.~H., and Yu, P. L.~H.
\newblock 2010.
\newblock Distance-based tree models for ranking data.
\newblock {\em Comput. Stat. Data Anal.} 54(6):1672--1682.

\bibitem[\protect\citeauthoryear{Lee and Yu}{2012}]{LeeMWD}
Lee, P., and Yu, P.
\newblock 2012.
\newblock Mixtures of weighted distance-based models for ranking data with
  applications in political studies.
\newblock 56:2486--2500.

\bibitem[\protect\citeauthoryear{Lundberg and Lee}{2017}]{LundbergL17}
Lundberg, S., and Lee, S.
\newblock 2017.
\newblock A unified approach to interpreting model predictions.
\newblock {\em CoRR} abs/1705.07874.

\bibitem[\protect\citeauthoryear{Montavon \bgroup et al\mbox.\egroup
  }{2015}]{MontavonBBSM15}
Montavon, G.; Bach, S.; Binder, A.; Samek, W.; and M{\"{u}}ller, K.
\newblock 2015.
\newblock Explaining nonlinear classification decisions with deep taylor
  decomposition.
\newblock {\em CoRR} abs/1512.02479.

\bibitem[\protect\citeauthoryear{Park and Jun}{2009}]{Park2009}
Park, H.-S., and Jun, C.-H.
\newblock 2009.
\newblock A simple and fast algorithm for k-medoids clustering.
\newblock {\em Expert Syst. Appl.} 36(2):3336--3341.

\bibitem[\protect\citeauthoryear{Ribeiro, Singh, and
  Guestrin}{2016}]{RibeiroSG16}
Ribeiro, M.~T.; Singh, S.; and Guestrin, C.
\newblock 2016.
\newblock "why should {I} trust you?": Explaining the predictions of any
  classifier.
\newblock {\em CoRR} abs/1602.04938.

\bibitem[\protect\citeauthoryear{Rousseeuw}{1987}]{rousseeuw1987silhouettes}
Rousseeuw, P.~J.
\newblock 1987.
\newblock Silhouettes: a graphical aid to the interpretation and validation of
  cluster analysis.
\newblock {\em Journal of computational and applied mathematics} 20:53--65.

\bibitem[\protect\citeauthoryear{Shieh, Bai, and Tsai}{2000}]{Shieh2000}
Shieh, G.~S.; Bai, Z.; and Tsai, W.-Y.
\newblock 2000.
\newblock Rank tests for independence — with a weighted contamination
  alternative.
\newblock {\em Statistica Sinica} 10(2):577--593.

\bibitem[\protect\citeauthoryear{Shrikumar, Greenside, and
  Kundaje}{2017}]{ShrikumarGK17}
Shrikumar, A.; Greenside, P.; and Kundaje, A.
\newblock 2017.
\newblock Learning important features through propagating activation
  differences.
\newblock {\em CoRR} abs/1704.02685.

\bibitem[\protect\citeauthoryear{Simonyan, Vedaldi, and
  Zisserman}{2013}]{SimonyanVZ13}
Simonyan, K.; Vedaldi, A.; and Zisserman, A.
\newblock 2013.
\newblock Deep inside convolutional networks: Visualising image classification
  models and saliency maps.
\newblock {\em CoRR} abs/1312.6034.

\bibitem[\protect\citeauthoryear{Sundararajan, Taly, and
  Yan}{2017}]{SundararajanTY17}
Sundararajan, M.; Taly, A.; and Yan, Q.
\newblock 2017.
\newblock Axiomatic attribution for deep networks.
\newblock {\em CoRR} abs/1703.01365.

\bibitem[\protect\citeauthoryear{Xu \bgroup et al\mbox.\egroup }{2013}]{XU2013}
Xu, W.; Hou, Y.; Hung, Y.; and Zou, Y.
\newblock 2013.
\newblock A comparative analysis of spearman's rho and kendall's tau in normal
  and contaminated normal models.
\newblock {\em Signal Processing} 93(1):261 -- 276.

\bibitem[\protect\citeauthoryear{Yang, Rangarajan, and
  Ranka}{2018}]{yang2018global}
Yang, C.; Rangarajan, A.; and Ranka, S.
\newblock 2018.
\newblock Global model interpretation via recursive partitioning.
\newblock {\em arXiv preprint arXiv:1802.04253}.

\end{thebibliography}

\section{Appendix}

\section{Background on Local Interpretability Techniques}
\subsection{Locally Interpretable Model-Agnostic Explanations (LIME)}

LIME produces interpretable representations of complicated models by optimizing two metrics: interpretability $\Omega$ and local fidelity $\mathcal{L}$. Defined as,

\begin{equation}
\zeta(x) ={\mathrm{argmin}}_{g \in G} \mathcal{L}(f, g, \pi_x) + \Omega(g).
\label{eq:lime}
\end{equation}

In Equation 1, the complexity or comprehensibility $\Omega$ of an interpretable function $g\in G$ (e.g. non-zero coefficients of a linear model) is optimized alongside the fidelity $\mathcal{L}$ or faithfulness of $g$ in approximating true function $f$ in local neighborhood $\pi_x$.

In practice, an input point $x$ is perturbed by random sampling in a local neighborhood and a simpler linear model is fit with the newly constructed synthetic data set. The method is model agnostic, which means it can be applied to neural networks or any other uninterpretable model. The now explainable linear model's weights can be used to interpret a particular model prediction.

\subsection{Integrated Gradients}

Sundararajan et. al developed an approach that satisfies two proposed axioms for attributions. First, sensitivity states that for any two inputs that differ in only a single feature and have different predictions, the attribution should be non-zero. Second, implementation invariance states that two networks that have equal outputs for all possible inputs should also have the same attributions. In this paper, the authors develop an approach that takes the path integral of the gradient for a particular point $x_i$ and the model's inference $F$ on the path ($\alpha$) between a zero information baseline $x'$ and the input $x$. 

$$IG_i(x) ::= (x_i - x'_i) \int_{\alpha=0}^1 \frac{\partial f(x'+\alpha \times (x-x'))}{\partial x_i}  d\alpha$$

The path integral between the baseline and the true input can be approximated with Riemman sums. An attribution vector for a particular observation (local) is produced as a result. The authors found that 20 to 300 steps can sufficiently approximate the integral within 5\%. Selecting a baseline vector remains an open research question. A domain-specific heuristic (e.g. a black image in image classification) has been recommended for baseline selection.

\subsection{Deep Learning Important FeaTures (DeepLIFT)}

Much like Integrated Gradients, DeepLIFT seeks to explain the difference in output $\Delta t$ between a 'reference'  or 'baseline' input $t^0$ and the original input $t$ for a neuron output: $\Delta t = t - t^0$. For each input $x_i$, an attribution score is calculated $C_{\Delta x_i \Delta t}$ that should sum up to the total change in the output $\Delta t$. DeepLIFT is best defined by this described summation-to-delta property: $$\sum_{i=1}^{n} C_{\Delta x_i \Delta t} = \Delta t$$

Shrikumar et. al then construct a function analogous to a partial derivative and use the chain rule to trace the attribution score from the output layer to the original input. This method also requires a reference vector and the authors recommend leveraging domain knowledge to select an optimal baseline input.

\section{Integrated Gradients and DeepLIFT Analysis for Synthetic Dataset}

While our local attributions using LIME did vary slightly depending our choice of kernel-width, 
we find local attributions to be extremely sensitive to our choice of a baseline input. Particularly, the number of points
in each cluster varies greatly depending on our choice of baseline for Integrated Gradients.
Both feature importances and subpopulations varied greatly for DeepLIFT.
Unfortunately in this context there is no natural choice of baselines. 
To select a baseline for Integrated Gradients, we ran a grid search across our range of inputs within +/-.005 to produce a neutral prediction.
We obtain two subpopulations using this baseline for Integrated Gradients. 
The first contains 610 samples and assigns an attribution of 89\% to feature A. 
The second contains the remaining 1390 and assigns a 99\% attribution to feature B.
We demonstrate that our global attribution technique is local-technique agnostic by generating new attributions using Integrated Gradients and DeepLIFT.

Both feature importances and subpopulations varied greatly for DeepLIFT.
To select a baseline for Integrated Gradients, we ran a grid search across our range of inputs within +/-.005 to produce a neutral prediction.
We obtain two subpopulation using this baseline for Integrated Gradients. 
The first contains 610 samples and assigns an attribution of 89\% to feature A. 
The second contains the remaining 1390 and assigns a 99\% attribution to feature B.

\section{GAM Applied to the Iris Dataset} 

We also evaluate the proposed global attribution method
on the well known Iris dataset (Anderson, 1936; Fisher, 1936).
The task is to classify Iris flowers into one of three species (setosa, virginica, and versicolor) 
based on the length and width of each sample's petals and sepals. 
We train a 2-hidden layer feed forward network on 75\% of the Iris data and validate on the remaining. 
We use a softmax activation function as the final output layer and ReLU activation functions for all other layers. 
The trained network achieves a validation accuracy of 90\%.

As with the synthetic dataset, we generate local attributions using LIME. We did not obtain
attributions using Integrated Gradients or DeepLIFT because that method relies on
the use of a baseline vector to determine the attributions, which is not as clearly defined for
the Iris dataset and defining one is beyond the scope of this paper.
We then conduct a silhouette analysis to determine the optimal number of clusters for our global attribution method. 
We find the peak silhouette score of 0.95 is achieved with three clusters, in contrast to 0.65 for two clusters and 0.78 for four clusters. 
We then apply our global attribution method using three subpopulations and extract three explainations to summarize the global attribution pattern.

\begin{figure}
    \includegraphics[width=\linewidth]{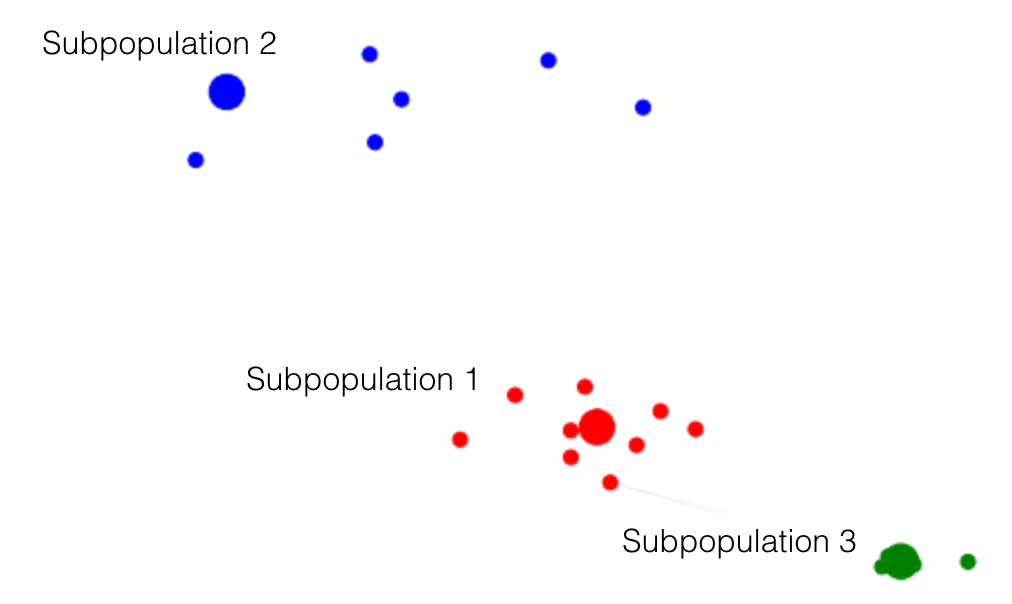}
\caption{
  \textbf{Attribution Subpopulations in Kendall Tau's Rank Distance Space.} 
    The attributions are represented as nodes in a force-directed graph where the distance between any two nodes is proportional to their Kendall Tau's rank distance. This allows us to visualize the relative rank distance among local attributions and their cluster medoids (represented as nodes with larger diameter). Subpopulation 1 and subpopulation 2 contain attributions associated with predictions for the setosa and versicolor flower species. Subpopulation 3 contains attributions associated with virginica predictions and a single versicolor prediction.}
\label{fig:irisClusters}
\end{figure}

To validate our results, we first plot the global attribution clusters in a force-directed graph to visualize Kendall Tau's rank distance among attributions. 
We find the global attributions form well-isolated clusters in rank distance space (see Figure ~\ref{fig:irisClusters}). 
Each cluster groups a set of similar attributions in rank space, which is succinctly summarized 
using each cluster's medoid.

We find each attribution cluster corresponds to a target class.
While this structure need not arise in the context of every problem, the attributions here can be explained nicely in terms of the 
classes associated with each prediction (a single exception being one versicolor prediction in subpopulation 3).
We did find slight variation in the assignment of nearby attribution vectors, namely those in the region between subpopulation 1 and subpopulation 3, 
but the cluster structure appeared stable across many iterations with distinct clusters.

\begin{center}
\begin{table}[h]
    \centering
\caption{Attribution Subpopulation Explanations \label{tab:irisMedoids} }
\begin{tabular}{ |cccc| }

 \hline
    Feature & Subpop. A  & Subpop. B & Subpop. C \\
 \hline

    $\begin{matrix} \text{sepal length} \\ \text{sepal width} \\ 
                     \text{petal length} \\ \text{petal width} \end{matrix}$ &
   $\begin{bmatrix} 0.01 \\ 0.61 \\ 0.03 \\ 0.35 \end{bmatrix}$ &
   $\begin{bmatrix} 0.15 \\ 0.31 \\ 0.37 \\ 0.17 \end{bmatrix}$ &
   $\begin{bmatrix} 0.11 \\ 0.01 \\ 0.40 \\ 0.48 \end{bmatrix}$ \\

 \hline
\end{tabular}
\end{table}
\end{center}

Each subpopulation's explanation succinctly summarizes the feature importances for each species' predictions (see Table ~\ref{tab:irisMedoids}). Sepal and petal width are most important for setosa predictions (subpopulation A), whereas sepal width and petal length are most important for versicolor predictions (subpopulation B). Petal features turn out to be most important for virginica predictions (subpopulation C). We can then compare these features against the weights of a baseline classifier to assess whether the medoids reasonably capture the correct set of feature importances.

\begin{center}
\begin{table}[h]
    \centering
\caption{Absolute Weights of Logistic Classifiers \label{tab:logisticWeights} }
\begin{tabular}{ |cccc| }

 \hline
    Feature & setosa  & versicolor & virginica \\
 \hline

    $\begin{matrix} \text{sepal length} \\ \text{sepal width} \\ 
                     \text{petal length} \\ \text{petal width} \end{matrix}$ &
   $\begin{bmatrix} 0.87 \\ 1.31 \\ 1.66 \\ 1.45 \end{bmatrix}$ &
   $\begin{bmatrix} 0.04 \\ 1.23 \\ 0.91 \\ 0.83 \end{bmatrix}$ &
   $\begin{bmatrix} 0.33 \\ 0.17 \\ 1.39 \\ 2.32 \end{bmatrix}$ \\

 \hline
\end{tabular}
\end{table}
\end{center}

We additionally validate our approach by comparing the set of feature importances from our global attributions against the weights learned from a 
baseline one-vs-rest logistic regression classifier.
We find the coefficients from the logistic regression classifiers correspond directly to the weights we obtain from our global attribution method.
We find sepal width, petal length, and petal weights to have the largest coefficients across the three classes, similar to
what we found across all three attribution medoids (see Table ~\ref{tab:logisticWeights}).  
Furthermore, we find the largest absolute coefficients across all classes (except for petal length for setosa) correspond directly to the feature 
importances uncovered by our global attribution method. 
For example, the largest weights for virginica predictions are the same petal features highlighted by the global attribution medoid for subpopulation C.

It is worth noting that two classifiers may learn a different set of feature interactions. In this case,  the feature interactions surfaced by our global attribution method correspond to the feature weights learned by a one-vs-rest logistic regression classifier. This provides evidence not only that the number of clusters and cluster patterns are meaningful, but also that the specific feature importances uncovered in each cluster effectively describe global model attribution.

\section{Mushroom Global Attributions}
Global Attributions with explanations for all features across the three techniques.

\subsection{Global Attributions with DeepLIFT}

\begin{figure}[H]
    \includegraphics[width=\linewidth]{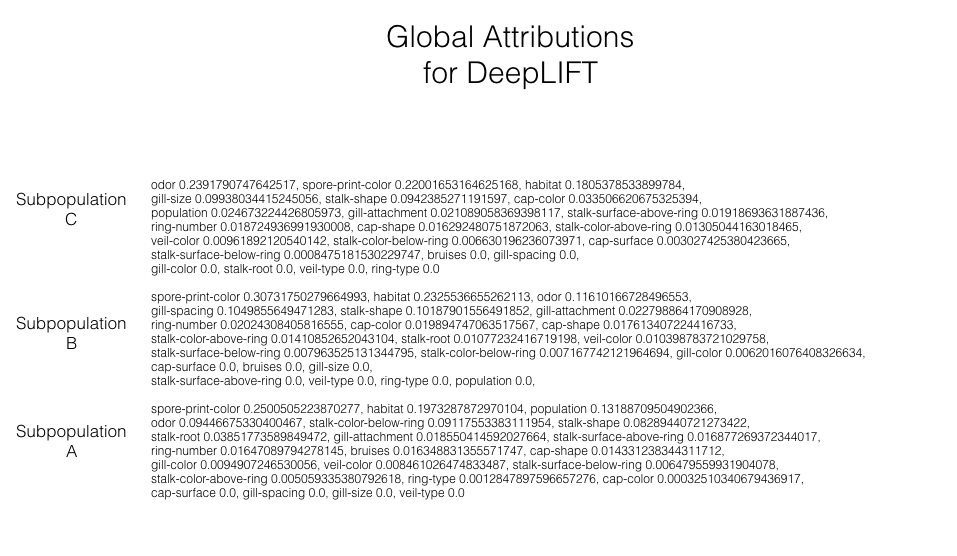}
\end{figure}

\subsection{Global Attributions with Integrated Gradients}

\begin{figure}[H]
    \includegraphics[width=\linewidth]{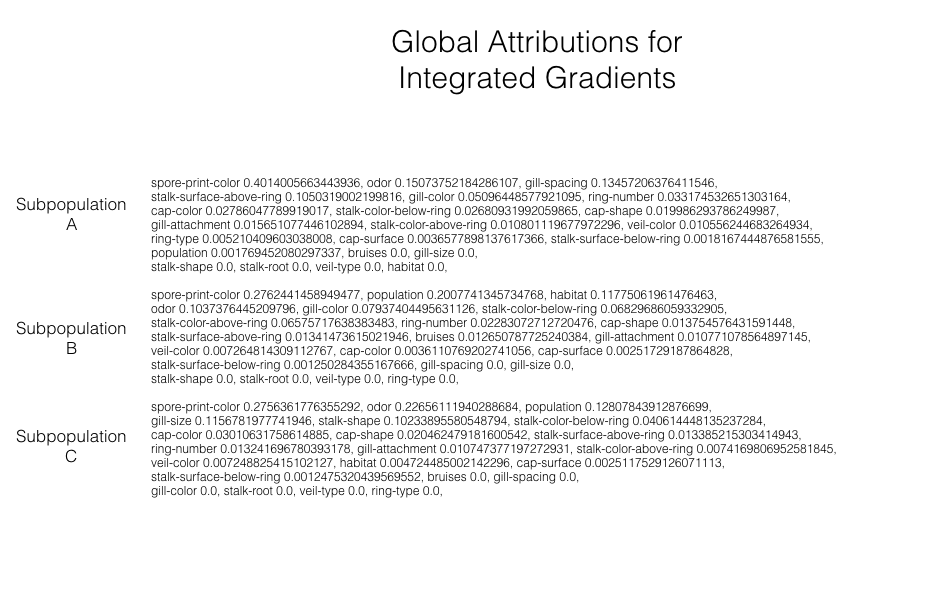}
\end{figure}

\subsection{Global Attributions with LIME}

\begin{figure}[H]
    \includegraphics[width=\linewidth]{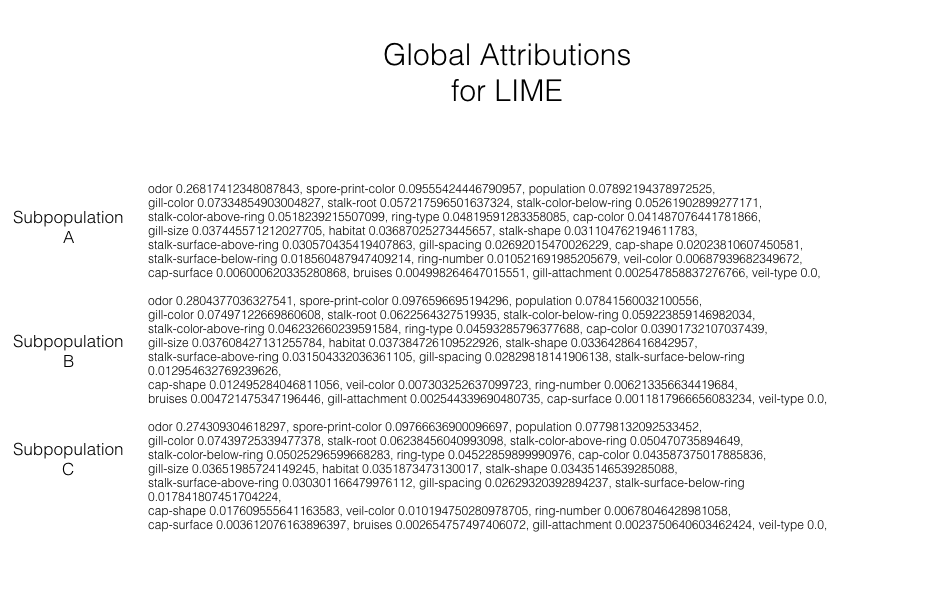}
\end{figure}

\section{User Studies}

We asked 55 MTurk users to identity the top five features based on GAM's global explanations with two subpopulations on the FICO dataset (see figures below). 
\begin{figure}[H]
    \includegraphics[width=\linewidth]{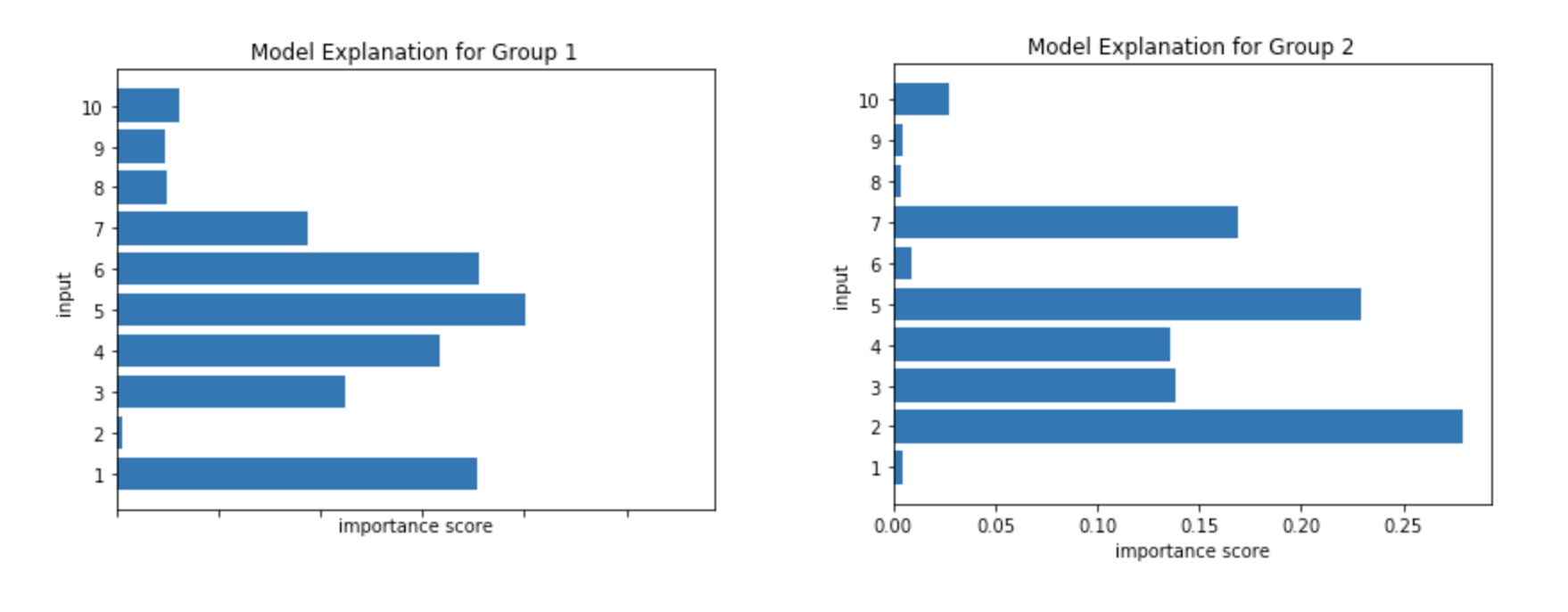}
\end{figure}

In 97\% of cases respondents selected the meaningful features for subpopulation one and 92\% for subpopulation two.

We also generated identical attributions for SP-LIME with B=2 and asked 55 credit modeling practitioners which explanation they preferred. 
In this case, the raw feature names were also displayed without a method label (see figures). Below are aggregate anonymous responses.

\begin{figure}[H]
    \includegraphics[width=\linewidth]{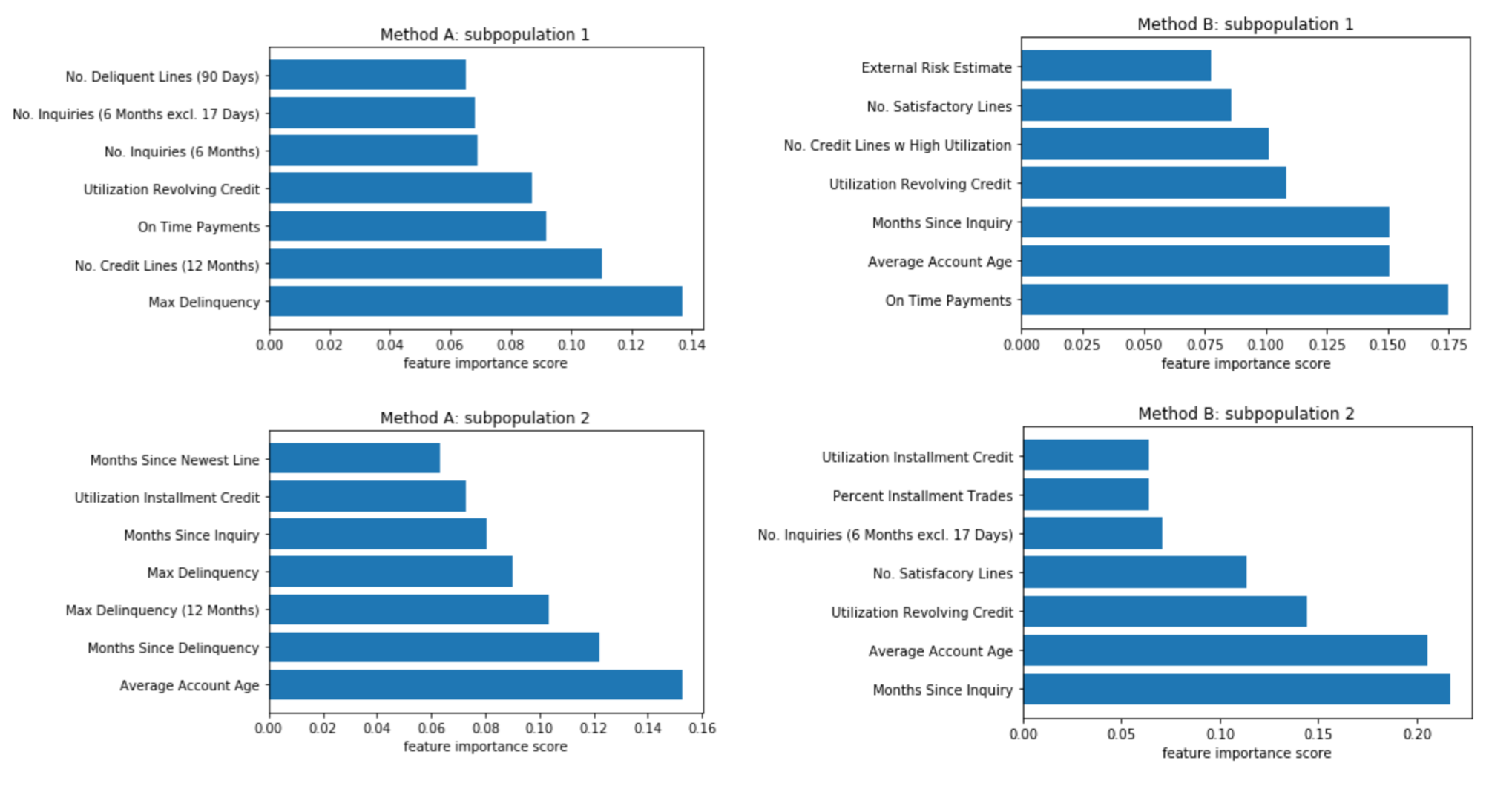}
\end{figure}

\begin{figure}[H]
    \includegraphics[width=\linewidth]{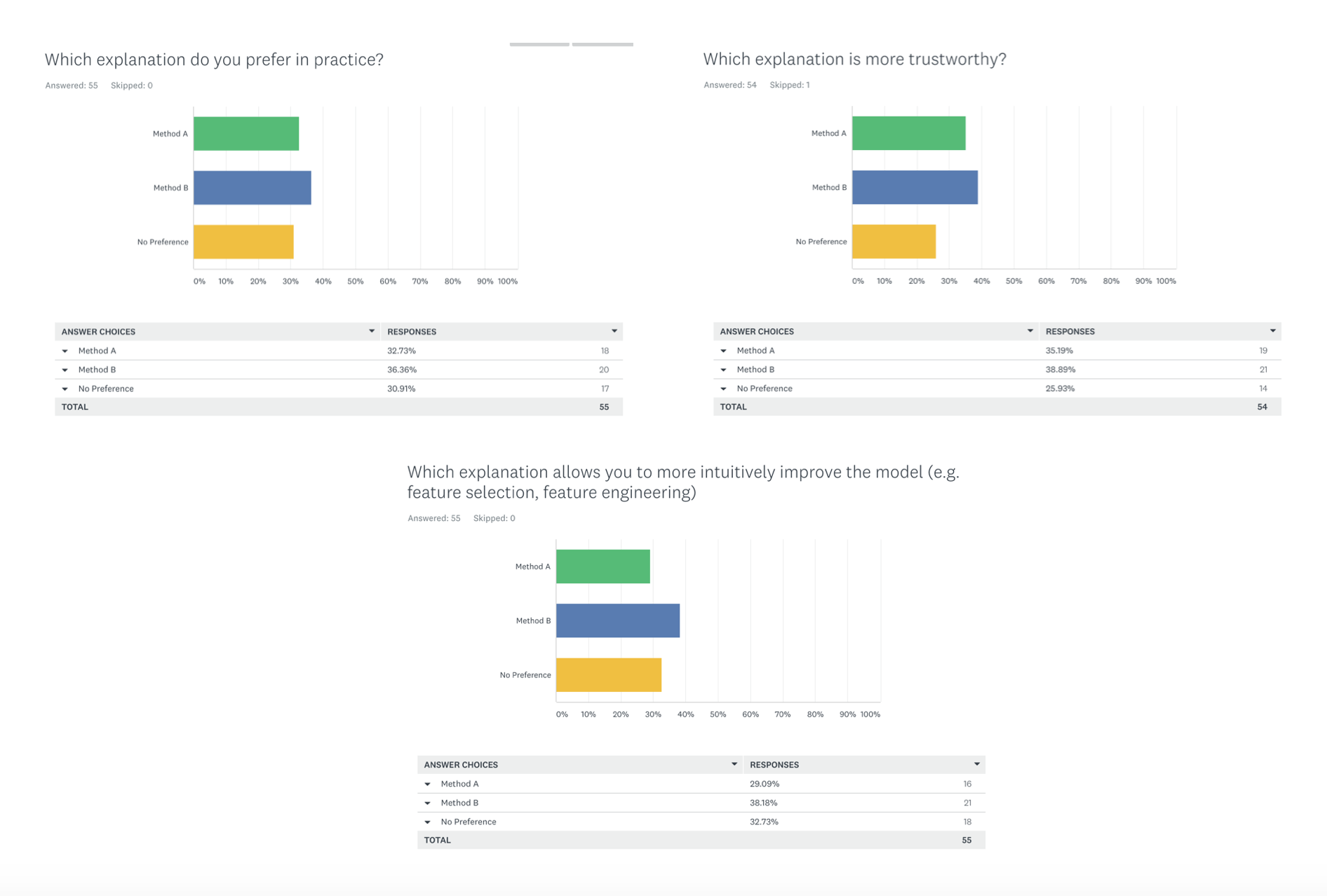}
\end{figure}

\end{document}